\documentclass[10pt,twocolumn,letterpaper]{article}

\usepackage{cvpr}
\usepackage{times}
\usepackage{epsfig}
\usepackage{graphicx}
\usepackage{amsmath}
\usepackage{amssymb}
\usepackage{multirow}
\usepackage{booktabs} 
\usepackage[]{algorithm2e}
\usepackage{algorithmic}
\usepackage{subcaption}
\usepackage{enumitem}

\def\ie{\emph{i.e}\onedot}
\def\eg{\emph{e.g}\onedot}

\def\jft{JFT dataset \cite{hinton2015distilling, chollet2017xception}\xspace}

\usepackage[pagebackref=true,breaklinks=true,letterpaper=true,colorlinks,bookmarks=false]{hyperref}

\cvprfinalcopy 

\def\cvprPaperID{4286} 

\ifcvprfinal\fi
\begin{document}

\title{Self-training with Noisy Student improves ImageNet classification}

\author{Qizhe Xie\thanks{~This work was conducted at Google.}~~$^1$, Minh-Thang Luong$^1$, Eduard Hovy$^2$, Quoc V. Le$^1$\\
$^1$Google Research, Brain Team, $^2$Carnegie Mellon University\\
{\tt\small \{qizhex, thangluong, qvl\}@google.com, hovy@cmu.edu}
}

\maketitle

\begin{abstract}
We present Noisy Student Training, a semi-supervised learning approach that works well even when labeled data is abundant.
Noisy Student Training  achieves 88.4\% top-1 accuracy on ImageNet, which is 2.0\% better than the state-of-the-art model that requires 3.5B weakly labeled Instagram images. On robustness test sets, it improves ImageNet-A top-1 accuracy from 61.0\% to 83.7\%, reduces ImageNet-C mean corruption error from 45.7 to 28.3, and reduces ImageNet-P mean flip rate from 27.8 to 12.2.

Noisy Student Training extends the idea of self-training and distillation with the use of equal-or-larger student models and noise added to the student during learning.
On ImageNet, we first train an EfficientNet model on labeled images and use it as a teacher to generate pseudo labels for 300M unlabeled images. We then train a larger EfficientNet as a student model on the combination of labeled and pseudo labeled images. We iterate this process by putting back the student as the teacher. During the learning of the student, we inject noise such as  dropout, stochastic depth, and data augmentation via RandAugment to the student so that the  student generalizes better than the teacher.\footnote{Models are available at \url{https://github.com/tensorflow/tpu/tree/master/models/official/efficientnet}. Code is available at \url{https://github.com/google-research/noisystudent}.}

\end{abstract}

\section{Introduction}

Deep learning has shown remarkable successes in image recognition in recent years~\cite{krizhevsky2012imagenet,szegedy2015going,simonyan2014very,he2016deep,tan2019efficientnet}. However state-of-the-art (SOTA) vision models are still trained with supervised learning which requires a large corpus of labeled images to work well. By  showing the models only labeled images, we limit ourselves from  making use of unlabeled images available in  much larger quantities to improve accuracy and robustness of SOTA models.

Here, we use unlabeled images to improve the SOTA ImageNet accuracy and show that the accuracy gain has an outsized impact on robustness (out-of-distribution generalization). For this purpose, we use a much larger corpus of unlabeled images, where a large fraction of images do not belong to ImageNet training set distribution (i.e., they do not belong to any category in ImageNet).
We train our model with Noisy Student Training, 
a semi-supervised learning approach,
which has three main steps: (1) train a teacher model on labeled images, (2) use the teacher to generate pseudo labels on unlabeled images, and (3) train a student model on the combination of labeled images and pseudo labeled images. We iterate this algorithm a few times by treating the student as a teacher to relabel the unlabeled data and training a new student.  

Noisy Student Training improves  self-training and distillation in two ways. First, it makes the student larger than, or at least equal to, the teacher so the student can better learn from a larger dataset. Second, it adds noise to the student so the noised student is forced to learn harder from the pseudo labels. To noise the student, we use input noise such as RandAugment data augmentation~\cite{cubuk2019randaugment} and model noise such as  dropout~\cite{srivastava2014dropout} and stochastic depth~\cite{huang2016deep} during training.

Using Noisy Student Training, together with 300M unlabeled images, we improve EfficientNet's~\cite{tan2019efficientnet} ImageNet top-1 accuracy to 88.4\%. 
This accuracy is 2.0\% better than the previous SOTA results which requires 3.5B weakly labeled Instagram images. Not only our method improves standard ImageNet  accuracy, it also  improves classification robustness on much harder test sets  by  large margins: ImageNet-A~\cite{hendrycks2019natural} top-1 accuracy from 61.0\% to 83.7\%, ImageNet-C~\cite{hendrycks2018benchmarking} mean corruption error (mCE) from 45.7 to 28.3 and ImageNet-P~\cite{hendrycks2018benchmarking} mean flip rate (mFR) from 27.8 to 12.2. Our main results are shown in Table~\ref{tab:summary}.

\begin{table}[h!]
\footnotesize
    \centering 
        \begin{tabular}{l|cccc}           
        \toprule 
          &  ImageNet & ImageNet-A & ImageNet-C & ImageNet-P \\
          &  top-1 acc. & top-1 acc. & mCE & mFR \\
        \midrule      
        
        Prev. SOTA  & 86.4\% & 61.0\% & 45.7 & 27.8\\  
        Ours & \bf 88.4\% & \bf 83.7\% & \bf 28.3 & \bf 12.2\\
        \bottomrule
        \end{tabular}
    \caption{Summary of key results compared to previous state-of-the-art models \cite{touvron2019fixing,mahajan2018exploring}. Lower is better for mean corruption error (mCE) and mean flip rate (mFR).}  
    \label{tab:summary}
\end{table}

\section{Noisy Student Training}
\label{sec:method}
\begin{figure}[h!]
    \centering
    \includegraphics[width=1.0\columnwidth]{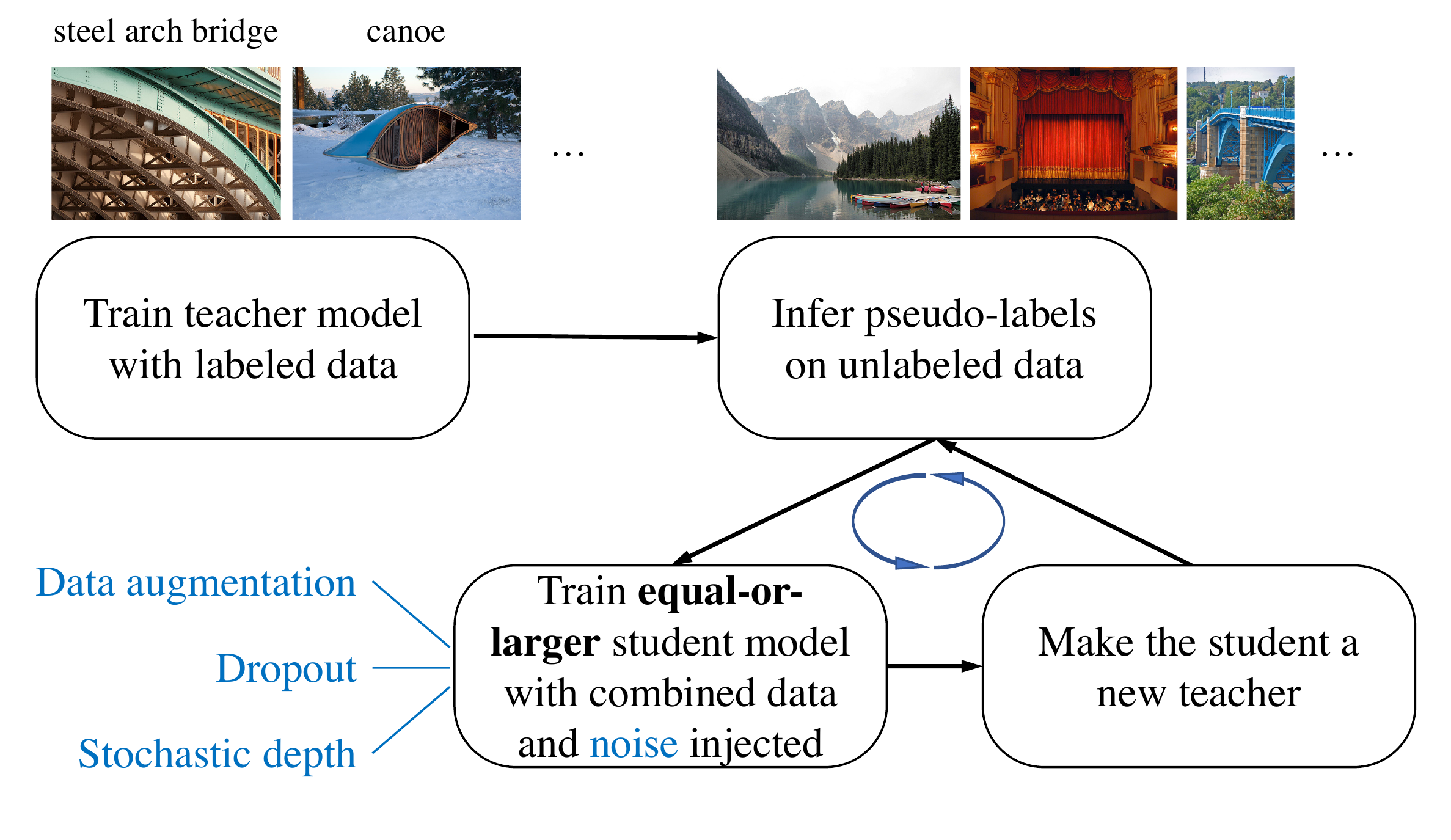}
    \caption{Illustration of the Noisy Student Training. (All shown images are from ImageNet.) }
    \label{fig:noisy_student}
\end{figure}

Algorithm~\ref{algo:noisy_student} gives an overview of  Noisy Student Training.
The inputs to the algorithm are both labeled and unlabeled images. We use the labeled images to train a teacher model using the standard cross entropy loss. We then use the teacher model to generate pseudo labels on unlabeled images. The pseudo labels can be soft (a continuous distribution) or hard (a one-hot distribution). We then train a student model which minimizes the combined cross entropy loss on both labeled images and unlabeled images. Finally, we iterate the process by putting back the student as a teacher to generate new pseudo labels and train a new student. The algorithm is also illustrated in Figure \ref{fig:noisy_student}.

\begin{algorithm}
\small
	\caption{Noisy Student Training.}
	\label{algo:noisy_student}
	\begin{algorithmic}[1]
		\REQUIRE 
		Labeled images $\{(x_1, y_1), (x_2, y_2), ..., (x_n, y_n)\}$ and unlabeled images $\{\tilde{x}_1, \tilde{x}_2, ..., \tilde{x}_m\}$.
		\STATE  Learn teacher model $\theta_{*}^t$ which minimizes the cross entropy loss on labeled images  $$\frac{1}{n}\sum_{i = 1}^n \ell(y_i, f^{noised}(x_i, \theta^t) )$$ \\ 
		\STATE Use a normal (i.e., not noised) teacher model to generate soft or hard pseudo labels for clean (i.e., not distorted) unlabeled images $$\tilde{y}_i = f(\tilde{x}_i, \theta_{*}^t), \forall i = 1, \cdots, m $$ \\
		\STATE Learn an {\bf equal-or-larger} student model $\theta_*^s$ which minimizes the cross entropy loss on labeled images and unlabeled images with {\bf noise} added to the student model $$\frac{1}{n}\sum_{i=1}^n \ell(y_i, f^{noised}(x_i, \theta^s) ) + \frac{1}{m} \sum_{i=1}^m \ell (\tilde{y}_i, f^{noised}(\tilde{x}_i, \theta^s)) $$\\
		\STATE Iterative training: Use the student as a teacher and go back to step 2.
	\end{algorithmic}
\end{algorithm}

 The algorithm is an improved version of self-training, a
 method in semi-supervised learning (\eg,~\cite{scudder1965probability,yarowsky1995unsupervised}), and distillation~\cite{hinton2015distilling}. 
More discussions on how our method is related to prior works are included in Section~\ref{sec:related_works}.

Our key improvements lie in adding noise to the student and using student models that are not smaller than the teacher. This makes our method different from Knowledge Distillation~\cite{hinton2015distilling} where 1) noise is often not used and 2) a smaller student model is often used to be faster than the teacher. One can think of our method as \emph{knowledge expansion} in which we want the student to be better than the teacher by giving the student model enough capacity and difficult environments in terms of noise to learn through.
 
{\bf Noising Student} -- When the student is deliberately noised it is trained to be consistent to the teacher that is not noised when it generates  pseudo labels.
In our experiments, we use two types of noise: input noise and model noise. For input noise, we use data augmentation with RandAugment~\cite{cubuk2019randaugment}. For model noise, we use dropout~\cite{srivastava2014dropout} and stochastic depth~\cite{huang2016deep}.

When applied to unlabeled data, noise has an important benefit of enforcing invariances in the decision function on both labeled and unlabeled data.
First, {\it data augmentation} is an important noising method in Noisy Student Training because it forces the student to ensure prediction consistency across augmented versions of an image (similar to UDA~\cite{uda}). Specifically, in our method, the teacher produces high-quality pseudo labels by reading in clean images, while the student is required to reproduce those labels with augmented images as input. For example, the student must ensure that a translated version of an image should have the same category as the original image. 
Second, when {\it dropout} and {\it stochastic depth function} are used as noise, the teacher behaves like an ensemble at inference time (when it generates pseudo labels), whereas the student behaves like a single model. In other words, the student is forced to mimic a more powerful ensemble model.
We present an ablation study on the effects of noise in Section \ref{sec:ablation_noise}.\looseness=-1

{\bf Other Techniques} -- Noisy Student Training also works better with an additional trick: data filtering and balancing, similar to~\cite{uda,billion_large_scale}.  
Specifically, we filter images that the teacher model has low confidences on since they are usually out-of-domain images. To ensure that the distribution of the unlabeled images match that of the training set, we also need to balance the number of unlabeled images for each class, as all classes in ImageNet have a similar number of labeled images. For this purpose, we duplicate images in classes where there are not enough images. For classes where we have too many images, we take the images with the highest confidence.\footnote{The benefits of data balancing is significant for small models while less significant for larger models. See Study \#5 in Appendix \ref{sec:ablation_study} for more details.}

Finally, we emphasize that our method can be used with soft or hard pseudo labels as both work well in our experiments. Soft pseudo labels, in particular, work slightly better for out-of-domain unlabeled data. Thus in the following, for consistency, we report results with soft pseudo labels unless otherwise indicated.  

\paragraph{Comparisons with Existing SSL Methods.}
Apart from self-training, another important line of work in semi-supervised learning~\cite{chapelle2009semi,zhu2005semi} is based on consistency training~\cite{bachman2014learning,rasmus2015semi,laine2016temporal,tarvainen2017mean,miyato2018virtual,uda,mixmatch} and pseudo labeling~\cite{lee2013pseudo,iscen2019label,shi2018transductive,arazo2019pseudo}. 
Although they have produced promising results, in our preliminary experiments, methods based on consistency regularization and pseudo labeling work less well on ImageNet. Instead of using a teacher model trained on labeled data to generate pseudo-labels, these methods do not have a separate teacher model and use the model being trained to generate pseudo-labels. In the early phase of training, the model being trained has low accuracy and high entropy, hence consistency training regularizes the model towards high entropy predictions, and prevents it from achieving good accuracy. 
A common workaround is to use entropy minimization, to filter examples with low confidence or to ramp up the consistency loss. However, the additional hyperparameters introduced by the ramping up schedule, confidence-based filtering and the entropy minimization make them more difficult to use at scale. The self-training / teacher-student framework is better suited for ImageNet because we can train a good teacher on ImageNet using labeled data.

\section{Experiments}
In this section, we will first describe our experiment details.
We will then present our ImageNet results compared with those of state-of-the-art models. Lastly, we demonstrate the surprising improvements of our models on robustness datasets (such as ImageNet-A, C and P) as well as under adversarial attacks.

\subsection{Experiment Details} 
\label{sec:exp_details}

\paragraph{Labeled dataset.} We conduct experiments on ImageNet 2012 ILSVRC challenge prediction task since it has been considered one of the most heavily benchmarked datasets in computer vision and that improvements on ImageNet transfer to other datasets \cite{kornblith2019better,recht2019imagenet}. 

\paragraph{Unlabeled dataset.} We obtain unlabeled images from the \jft, which has around 300M images. Although the images in the dataset have labels, we ignore the labels and treat them as unlabeled data. We filter the ImageNet validation set images from the dataset (see~\cite{ngiam2018domain}). 

We then perform data filtering and balancing on this corpus. First, we run an EfficientNet-B0 trained on ImageNet~\cite{tan2019efficientnet} over the \jft to predict a label for each image. We then select images that have confidence of the label higher than 0.3.  For each class, we select at most 130K images that have the highest confidence. Finally, for classes that have less than 130K images, we duplicate some images at random so that each class can have 130K images. Hence the total number of images that we use for training a student model is 130M (with some duplicated images). Due to duplications, there are only 81M unique images among these 130M images. We do not tune these hyperparameters extensively since our method is highly robust to them.

To enable fair comparisons with our results, we also experiment with a public dataset YFCC100M~\cite{thomee2016yfcc100m} and show the results in Appendix~\ref{sec:exp_yfcc100m}.

\paragraph{Architecture.} We use EfficientNets~\cite{tan2019efficientnet} as our baseline models because they provide better capacity for more data. In our experiments, we also further scale up EfficientNet-B7 and obtain EfficientNet-L2. EfficientNet-L2 is
wider and deeper than EfficientNet-B7 but uses a lower resolution, which gives it more parameters to fit a large number of unlabeled images. Due to the large model size, the training time of EfficientNet-L2 is approximately five times the training time of EfficientNet-B7.  For more information about EfficientNet-L2, please refer to Table~\ref{tab:efficientnet_architectures} in Appendix~\ref{sec:architecture_details}.

\paragraph{Training details.} For labeled images, we use a batch size of 2048 by default and reduce the batch size when we could not fit the model into the memory. We find that using a batch size of 512, 1024, and 2048 leads to the same performance. We determine the number of training steps and the learning rate schedule by the batch size for labeled images. Specifically, we train the student model for 350 epochs for models larger than EfficientNet-B4, including EfficientNet-L2 and train smaller student models for 700 epochs. The learning rate starts at 0.128 for labeled batch size 2048 and decays by 0.97 every 2.4 epochs if trained for 350 epochs or every 4.8 epochs if trained for 700 epochs. 

We use a large batch size for unlabeled images, especially for large models, to make full use of large quantities of  unlabeled images. Labeled images and unlabeled images are concatenated together to compute the average cross entropy loss.
We apply the recently proposed technique to fix train-test resolution discrepancy~\cite{touvron2019fixing} for EfficientNet-L2. We first perform normal training with a smaller resolution for 350 epochs. Then we finetune the model with a larger resolution for 1.5 epochs on unaugmented labeled images. Similar to~\cite{touvron2019fixing}, we fix the shallow layers during finetuning.

Our largest model, EfficientNet-L2, needs to be trained for 6 days on a Cloud TPU v3 Pod, which has 2048 cores, if the unlabeled batch size is 14x the labeled batch size.

\begin{table*}[h!]
\small
    \centering      
        \begin{tabular}{l|cc|cc}                                                
        \toprule 
        Method & \# Params & Extra Data  &  Top-1 Acc. & Top-5 Acc.   \\
        \midrule       
        ResNet-50 \cite{he2016deep} & 26M & -  & 76.0\% & 93.0\%   \\
        ResNet-152 \cite{he2016deep} & 60M  & - &  77.8\% & 93.8\% \\
        DenseNet-264 \cite{huang2017densely} & 34M & - & 77.9\% & 93.9\%   \\
        Inception-v3 \cite{szegedy2016rethinking}& 24M  & - & 78.8\% & 94.4\%  \\
        Xception \cite{chollet2017xception} & 23M & - & 79.0\%   & 94.5\%  \\
        Inception-v4 \cite{szegedy2017inception} & 48M  & - & 80.0\% & 95.0\%  \\  
        Inception-resnet-v2  \cite{szegedy2017inception} & 56M & - & 80.1\% & 95.1\% \\  
		ResNeXt-101 \cite{xie2017aggregated}  & 84M   & - & 80.9\% & 95.6\% \\
		PolyNet \cite{zhang2017polynet}  & 92M  & - & 81.3\%  & 95.8\% \\
		SENet \cite{hu2018squeeze} & 146M & - & 82.7\% & 96.2\%   \\
		 NASNet-A \cite{zoph2018learning} & 89M & - & 82.7\% & 96.2\%   \\
		 AmoebaNet-A \cite{real2019regularized} & 87M & - & 82.8\% & 96.1\%   \\
		 PNASNet \cite{liu2018progressive} & 86M  & - & 82.9\% & 96.2\%  \\
		 AmoebaNet-C  \cite{cubuk2018autoaugment}  & 155M  & - &  83.5\%  & 96.5\% \\
		GPipe \cite{gpipe18} & 557M & - & 84.3\%  & 97.0\%   \\
		 EfficientNet-B7 \cite{tan2019efficientnet} &  66M  & - & 85.0\% & 97.2\%   \\
		 EfficientNet-L2 \cite{tan2019efficientnet} &  480M  & - & 85.5\% & 97.5\%   \\
        \midrule
        ResNet-50 Billion-scale \cite{billion_large_scale} & 26M & \multirow{4}{*}{3.5B images labeled with tags} & 81.2\% & 96.0\% \\
        ResNeXt-101 Billion-scale \cite{billion_large_scale} & 193M &  & 84.8\% & - \\
        ResNeXt-101 WSL \cite{mahajan2018exploring} & 829M &   & 85.4\% & 97.6\% \\
         FixRes ResNeXt-101 WSL \cite{touvron2019fixing} & 829M  &  & 86.4\% &  98.0\%  \\
         \midrule
         Big Transfer (BiT-L) \cite{kolesnikov2019large}$^{\dagger}$ & 928M  & 300M weakly labeled images from JFT & 87.5\% &  98.5\%  \\
         \midrule
         \bf Noisy Student Training (EfficientNet-L2) &  480M & 300M unlabeled images from JFT & \bf 88.4\% & \bf 98.7\%  \\
        \bottomrule
        \end{tabular}
    \caption{Top-1 and Top-5 Accuracy of Noisy Student Training and  previous state-of-the-art methods on ImageNet. EfficientNet-L2  with Noisy Student Training is the result of iterative training for multiple iterations by putting back the student model as the new teacher. It has better tradeoff in terms of accuracy and model size compared to previous state-of-the-art models. $^{\dagger}$: Big Transfer is a concurrent work that performs transfer learning from the JFT dataset. 
    }  
    \label{tab:imagenet}
\end{table*}

\paragraph{Noise.} We use stochastic depth~\cite{huang2016deep}, dropout~\cite{srivastava2014dropout}, and RandAugment~\cite{cubuk2019randaugment} to noise the student. The hyperparameters for these noise functions are the same for EfficientNet-B7 and L2. In particular, we set the survival probability in stochastic depth to 0.8 for the final layer and follow the linear decay rule for other layers. We apply dropout to the final  layer with a dropout rate of 0.5. For RandAugment, we apply two random operations with magnitude set to 27.

\paragraph{Iterative training.} The best model in our experiments is a result of three iterations of putting back the student as the new teacher.
We first trained an EfficientNet-B7 on ImageNet as the teacher model. Then by using the B7 model as the teacher, we trained an EfficientNet-L2 model with the unlabeled batch size set to 14 times the labeled batch size. Then, we trained a new EfficientNet-L2 model with the EfficientNet-L2 model as the teacher. Lastly, we iterated again and used an unlabeled batch size of 28 times the labeled batch size. The detailed results of the three iterations are available in Section \ref{sec:iterative_training}.

\subsection{ImageNet Results}
\label{sec:imagenet_results}
We first report the validation set accuracy on the ImageNet  2012 ILSVRC challenge prediction task as commonly done in literature~\cite{krizhevsky2012imagenet,szegedy2015going,he2016deep,tan2019efficientnet} (see also \cite{recht2019imagenet}). As shown in Table~\ref{tab:imagenet}, EfficientNet-L2 with Noisy Student Training achieves 88.4\% top-1 accuracy  which is significantly better than the best reported accuracy on EfficientNet of 85.0\%. The total gain of 3.4\% comes from two sources: by making the model larger (+0.5\%) and by Noisy Student Training (+2.9\%). In other words, Noisy Student Training makes a much larger impact on the accuracy than changing the architecture. 

Further, Noisy  Student Training outperforms the state-of-the-art accuracy of 86.4\% by FixRes ResNeXt-101 WSL~\cite{mahajan2018exploring,touvron2019fixing} that requires 3.5 Billion Instagram images labeled with tags. As a comparison, our method only requires 300M unlabeled images, which is perhaps more easy to collect. Our model is also approximately twice as small in the number of parameters compared to FixRes ResNeXt-101 WSL.

\begin{figure}[h!]
    \centering
    \includegraphics[width=0.95\columnwidth]{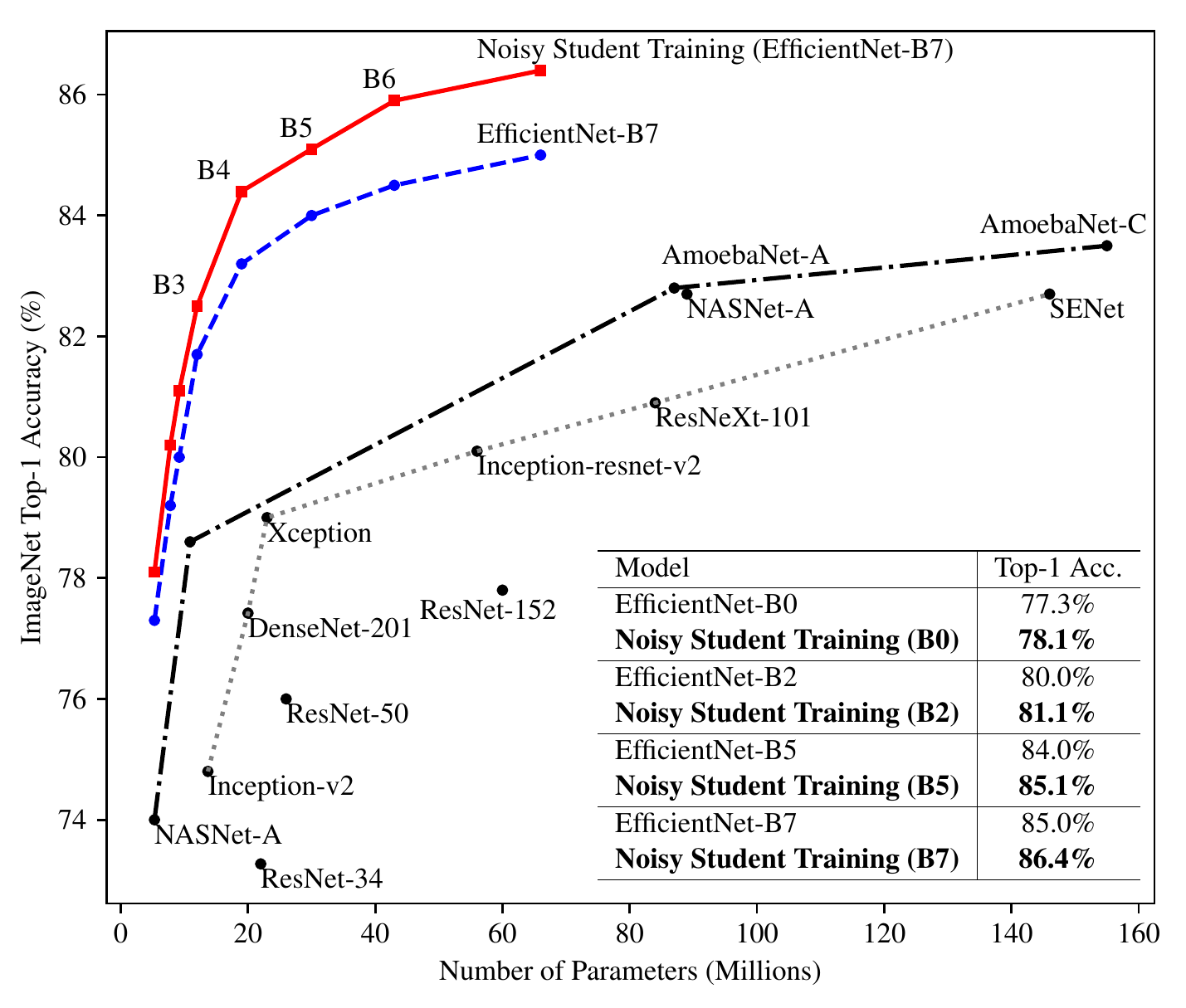}
    \caption{Noisy Student Training leads to significant improvements across all model sizes. We use the same architecture for the teacher and the student and do not perform iterative training.}
    \label{fig:vary_model_size}
\end{figure}

\paragraph{Model size study: Noisy Student Training for EfficientNet B0-B7 without Iterative Training.}  
 In addition to improving state-of-the-art results, we conduct experiments to verify if Noisy Student Training can benefit other EfficienetNet models. In previous experiments, iterative training was used to optimize the accuracy of EfficientNet-L2 but here we skip it as it is difficult to use iterative training for many experiments.
We vary the model size from EfficientNet-B0 to EfficientNet-B7~\cite{tan2019efficientnet} and use the same model as both the teacher and the student. We apply RandAugment to all EfficientNet baselines, leading to more competitive baselines. We set the unlabeled batch size to be three times the batch size of labeled images for all model sizes except for EfficientNet-B0. For EfficientNet-B0, we set the unlabeled batch size to be the same as the batch size of labeled images.
As shown in Figure \ref{fig:vary_model_size}, Noisy Student Training leads to a consistent improvement of around 0.8\% for all model sizes.  Overall, EfficientNets with Noisy Student Training provide a much better tradeoff between model size and accuracy than prior works. The results also confirm that  vision models can benefit from Noisy Student Training even without iterative training.

\subsection{Robustness Results on ImageNet-A, ImageNet-C and ImageNet-P}

We evaluate the best model, that achieves 88.4\% top-1 accuracy, on three robustness test sets: ImageNet-A, ImageNet-C and ImageNet-P. ImageNet-C and P test sets~\cite{hendrycks2018benchmarking} include images with common corruptions and perturbations such as blurring, fogging, rotation and scaling. ImageNet-A test set~\cite{hendrycks2019natural} consists of difficult images that cause significant drops in accuracy to state-of-the-art models. These test sets are considered as ``robustness'' benchmarks because the test images are either much harder, for ImageNet-A, or the test images are different from the training images, for ImageNet-C and P.

\begin{table}[h!]
\small
    \centering 
        \begin{tabular}{l|ccc}                                                
        \toprule 
        Method &  Top-1 Acc. & Top-5 Acc. \\
        \midrule      
        ResNet-101~\cite{hendrycks2019natural} & 4.7\% & - \\
        ResNeXt-101~\cite{hendrycks2019natural} (32x4d) & 5.9\% & -  \\
        ResNet-152~\cite{hendrycks2019natural} & 6.1\% & - \\
        ResNeXt-101~\cite{hendrycks2019natural} (64x4d) & 7.3\% & - \\
        DPN-98~\cite{hendrycks2019natural} & 9.4\% & - \\
        ResNeXt-101+SE~\cite{hendrycks2019natural} (32x4d) &  14.2\% & - \\
        ResNeXt-101 WSL~\cite{mahajan2018exploring,orhan2019robustness} & 61.0\% & - \\
        \midrule
	    EfficientNet-L2  &  49.6\% & 78.6\%   \\
		\bf Noisy Student Training (L2) & \bf 83.7\% & \bf 95.2\%  \\
        \bottomrule
        \end{tabular}
    \caption{Robustness results on ImageNet-A.}  
    \label{tab:robustness1}
\end{table}

\begin{table}[h!]
\small
    \centering 
        \begin{tabular}{l|c|cc}         
        \toprule 
        Method & Res. & Top-1 Acc. & mCE \\
        \midrule
        ResNet-50 \cite{hendrycks2018benchmarking} & 224 & 39.0\% & 76.7 \\
        SIN \cite{geirhos2018imagenet} &224 & 45.2\% & 69.3 \\
        Patch Gaussian \cite{lopes2019improving}  & 299 & 52.3\% & 60.4 \\
        ResNeXt-101 WSL~\cite{mahajan2018exploring,orhan2019robustness} & 224 & - & 45.7 \\
        \midrule
	    EfficientNet-L2  & 224 &  62.6\% &  47.5 \\
		Noisy Student Training (L2) & 224  &  76.5\% &  30.0 \\
		
	    EfficientNet-L2  & 299 & 66.6\% & 42.5  \\
		\bf Noisy Student Training (L2) & 299 & \bf 77.8\% & \bf 28.3 \\ 
        \bottomrule
        \end{tabular}
    \caption{Robustness results on ImageNet-C. mCE is the weighted average of error rate on different corruptions, with AlexNet's error rate as a baseline (lower is better).
    }  
    \label{tab:robustness2}
\end{table}

\begin{table}[h!]
\small
    \centering
        \begin{tabular}{l|c|cc}         
        \toprule 
        Method & Res. & Top-1 Acc. & mFR \\
        \midrule
        ResNet-50 \cite{hendrycks2018benchmarking} & 224 & - & 58.0 \\
        Low Pass Filter Pooling \cite{zhang2019making} &224 & - & 51.2 \\
        
        ResNeXt-101 WSL~\cite{mahajan2018exploring,orhan2019robustness} & 224 & - & 27.8  \\
        \midrule
	    EfficientNet-L2  & 224 & 80.4\% & 27.2  \\
		Noisy Student Training (L2) & 224 & 85.2\% & 14.2  \\ 
	    EfficientNet-L2  & 299 & 81.6\% &  23.7 \\
		\bf Noisy Student Training (L2) & 299 & \bf 86.4\% & \bf 12.2 \\
        \bottomrule
        \end{tabular}
    \caption{Robustness results on ImageNet-P, where images are generated with a sequence of perturbations. mFR measures the model's probability of flipping predictions under perturbations with AlexNet as a baseline (lower is better).
    }  
    \label{tab:robustness3}
\end{table}

\begin{figure*}[!htb]
\begin{subfigure}{.33\textwidth}
\centering
  \includegraphics[width=\linewidth]{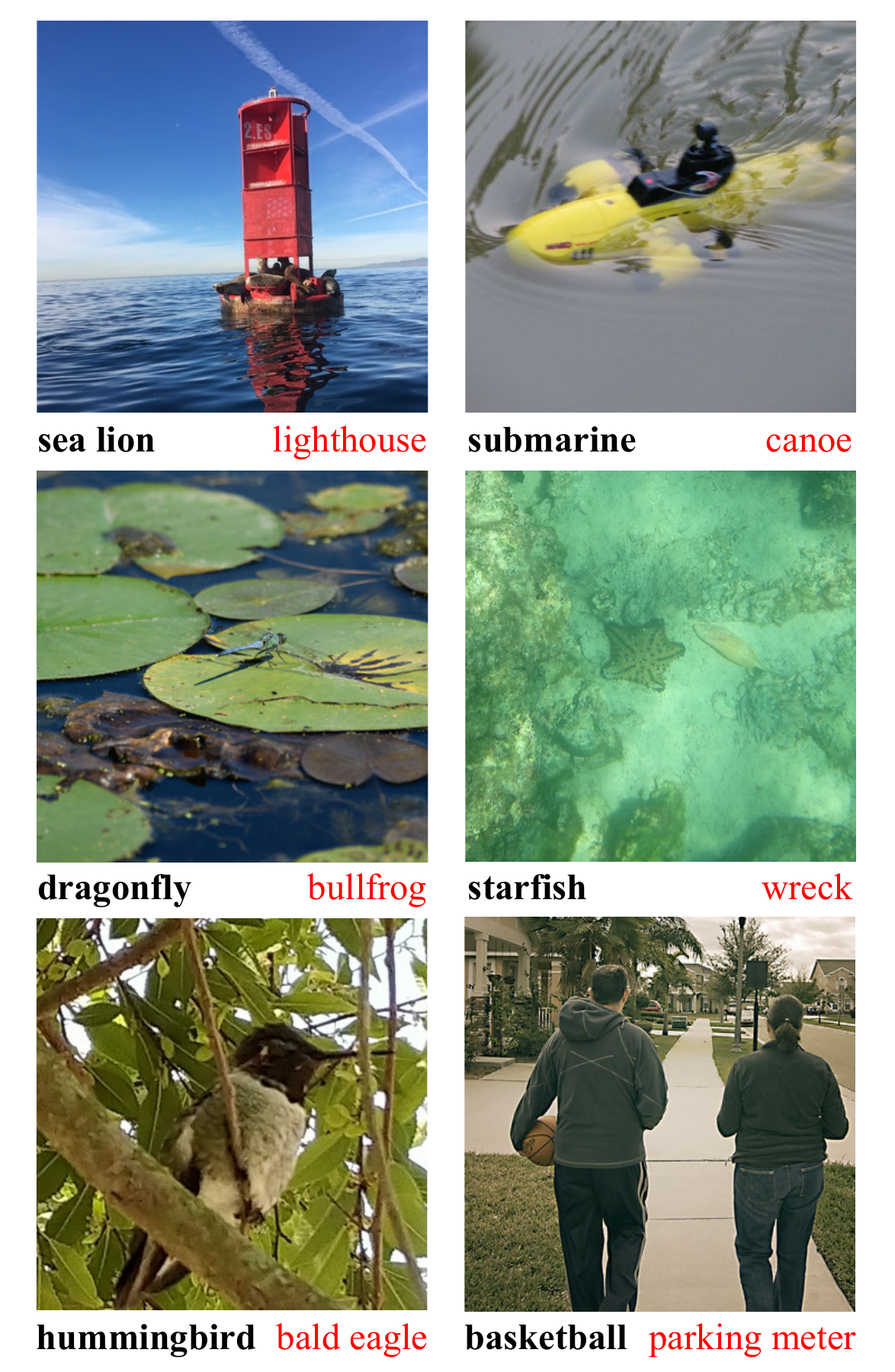}
    \caption{ImageNet-A}
        \label{fig:imagenet_a}
\end{subfigure}\hfill
\begin{subfigure}{.33\textwidth}
  \includegraphics[width=\linewidth]{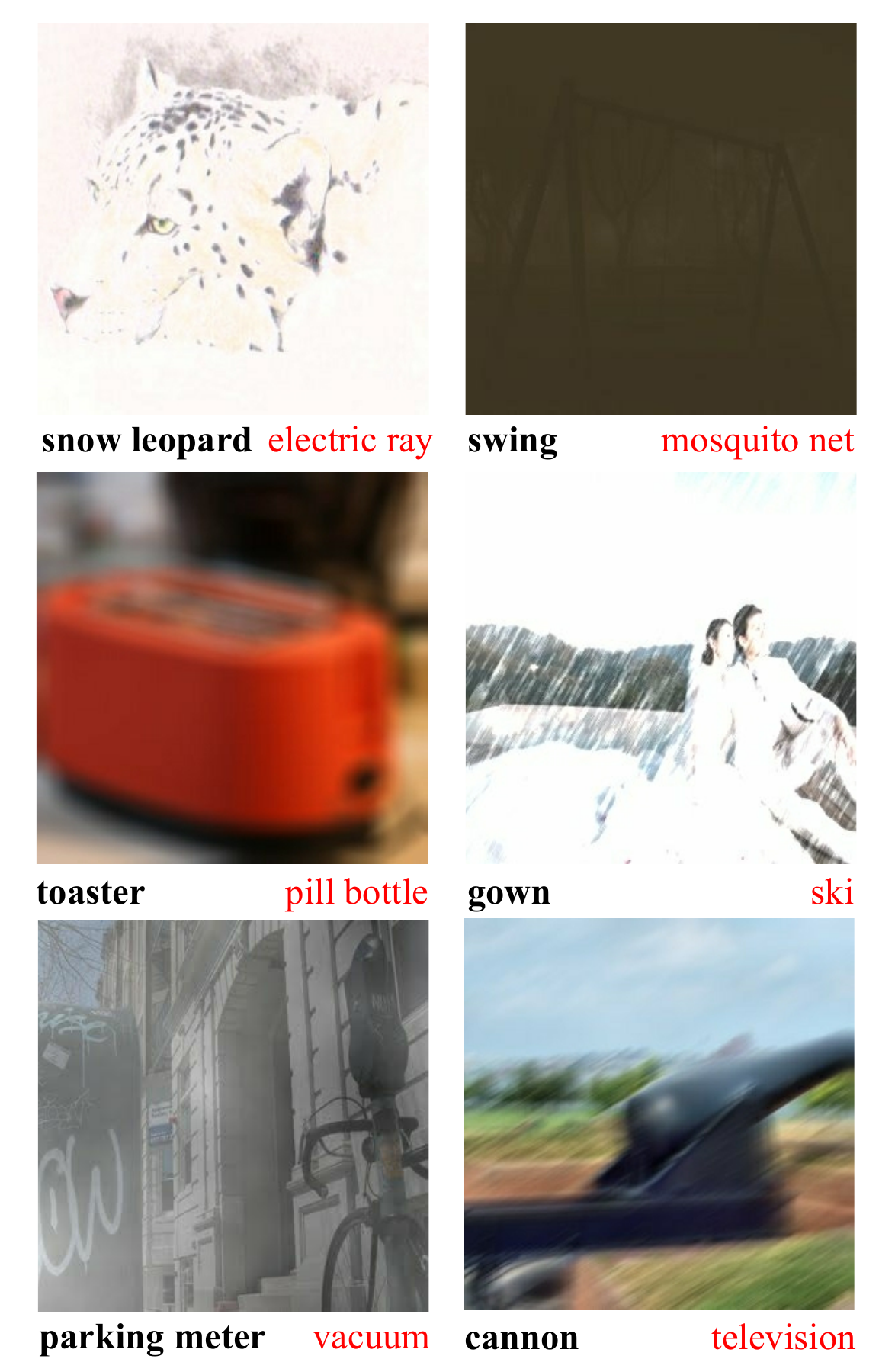}
\caption{ImageNet-C}
    \label{fig:imagenet_c}
\end{subfigure}\hfill
\begin{subfigure}{.33\textwidth}
  \includegraphics[width=\linewidth]{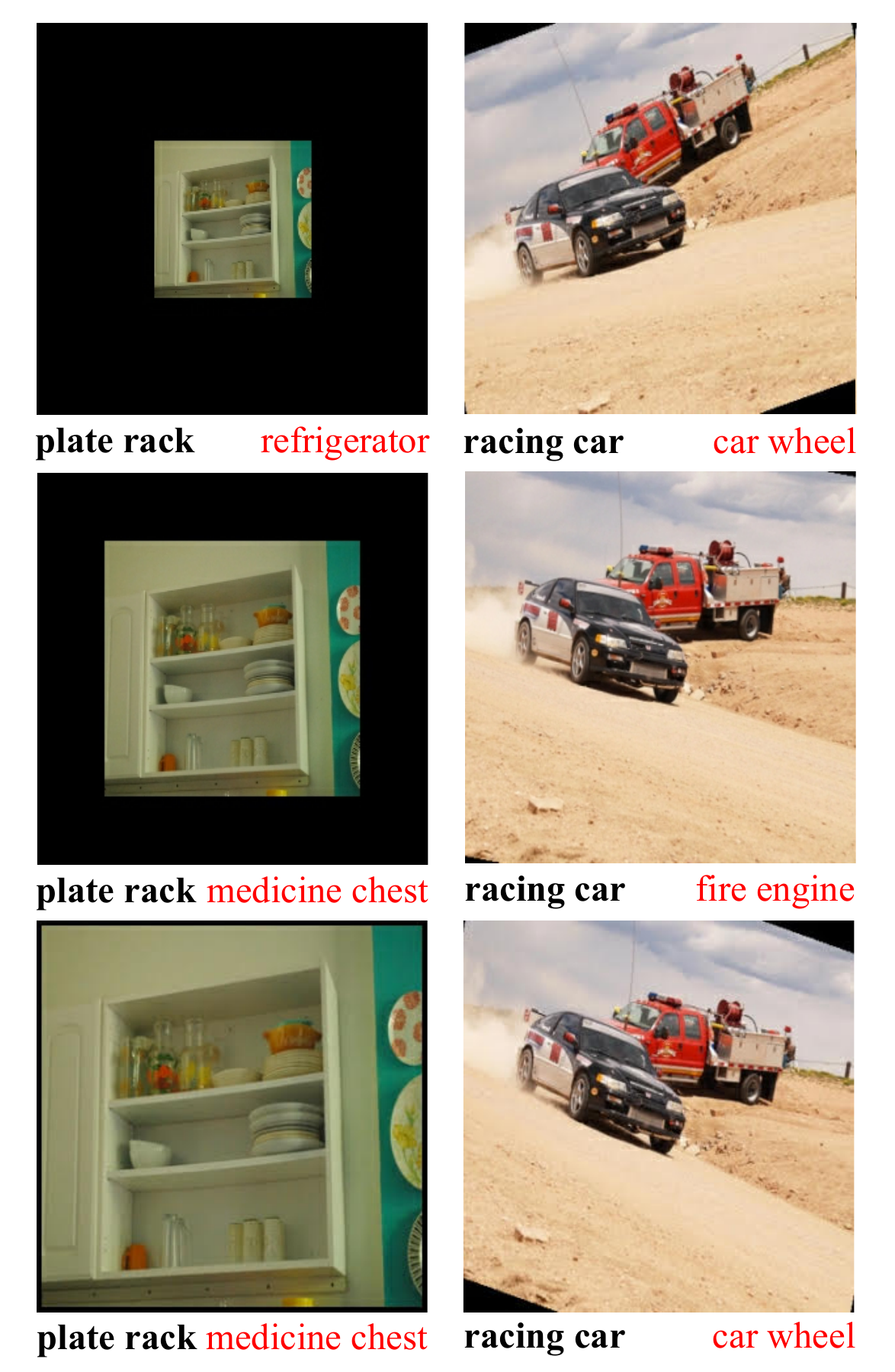}
\caption{ImageNet-P}
\label{fig:imagenet_p}
\end{subfigure}\hfill
  \caption{Selected images from robustness benchmarks ImageNet-A, C and P. Test images from ImageNet-C underwent artificial transformations (also known as common corruptions) that cannot be found on the ImageNet training set. Test images on ImageNet-P underwent different scales of perturbations. On ImageNet-A, C, EfficientNet with Noisy Student Tranining produces correct top-1 predictions (shown in \textbf{bold black} texts) and EfficientNet without Noisy Student Training produces incorrect top-1 predictions (shown in \textcolor{red}{red} texts). On ImageNet-P, EfficientNet without Noisy Student Training flips predictions frequently.}
\label{fig:robustnessacp}
\end{figure*}

For ImageNet-C and ImageNet-P, we evaluate models on two released versions with resolution 224x224 and 299x299 and resize images to the resolution EfficientNet trained on. 
As shown in Table~\ref{tab:robustness1},~\ref{tab:robustness2} and~\ref{tab:robustness3}, Noisy Student Training yields substantial gains on robustness datasets compared to the previous state-of-the-art model ResNeXt-101 WSL~\cite{mahajan2018exploring,orhan2019robustness} trained on 3.5B weakly labeled images. On ImageNet-A, it improves the top-1 accuracy from 61.0\% to 83.7\%. On ImageNet-C, it reduces mean corruption error (mCE) from 45.7 to 28.3. On ImageNet-P, it leads to a mean flip rate (mFR) of 14.2 if we use a resolution of 224x224 (direct comparison) and 12.2 if we use a resolution of 299x299.\footnote{For EfficientNet-L2, we use the model without finetuning with a larger test time resolution, since a larger resolution results in a discrepancy with the resolution of data and leads to degraded performance on ImageNet-C and ImageNet-P.}
These significant gains in robustness in ImageNet-C and ImageNet-P are surprising because our method was not deliberately optimized for robustness.\footnote{Note that both our model and ResNeXt-101 WSL use augmentations that have a small overlap with corruptions in ImageNet-C, which might result in better performance. 
Specifically, RandAugment includes augmentation Brightness, Contrast and Sharpness. ResNeXt-101 WSL uses augmentation of Brightness and Contrast.}

\paragraph{Qualitative Analysis.}

To intuitively understand the significant improvements on the three robustness benchmarks, we show several images in Figure~\ref{fig:robustnessacp} where the predictions of the standard model are incorrect while the predictions of the model with Noisy Student Training are correct. 

Figure \ref{fig:imagenet_a} shows example images from ImageNet-A and the predictions of our models. The model with Noisy Student Training can successfully predict the correct labels of these highly difficult images. For example, without Noisy Student Training, the model predicts \emph{bullfrog} for the image shown on the left of the second row, which might be resulted from the black lotus leaf on the water. With Noisy Student Training, the model correctly predicts \emph{dragonfly} for the image. At the top-left image, the model without Noisy Student Training ignores the \emph{sea lion}s and mistakenly recognizes a buoy as a lighthouse, while the model with Noisy Student Training can recognize the \emph{sea lion}s.

Figure \ref{fig:imagenet_c} shows images from ImageNet-C and the corresponding predictions. As can be seen from the figure, our model with Noisy Student Training makes correct predictions for images under severe corruptions and perturbations such as snow, motion blur and fog, while the model without Noisy Student Training suffers greatly under these conditions. The most interesting image is shown on the right of the first row. The \emph{swing} in the picture is barely recognizable by human while the model with Noisy Student Training still makes the correct prediction.

Figure \ref{fig:imagenet_p} shows images from ImageNet-P and the corresponding predictions. As can be seen, our model with Noisy Student Training makes correct and consistent predictions as images undergone different  perturbations while the model without Noisy Student Training flips predictions frequently.

\subsection{Adversarial Robustness Results}

\begin{figure}[h!]
\centering
  \includegraphics[width=0.85\columnwidth]{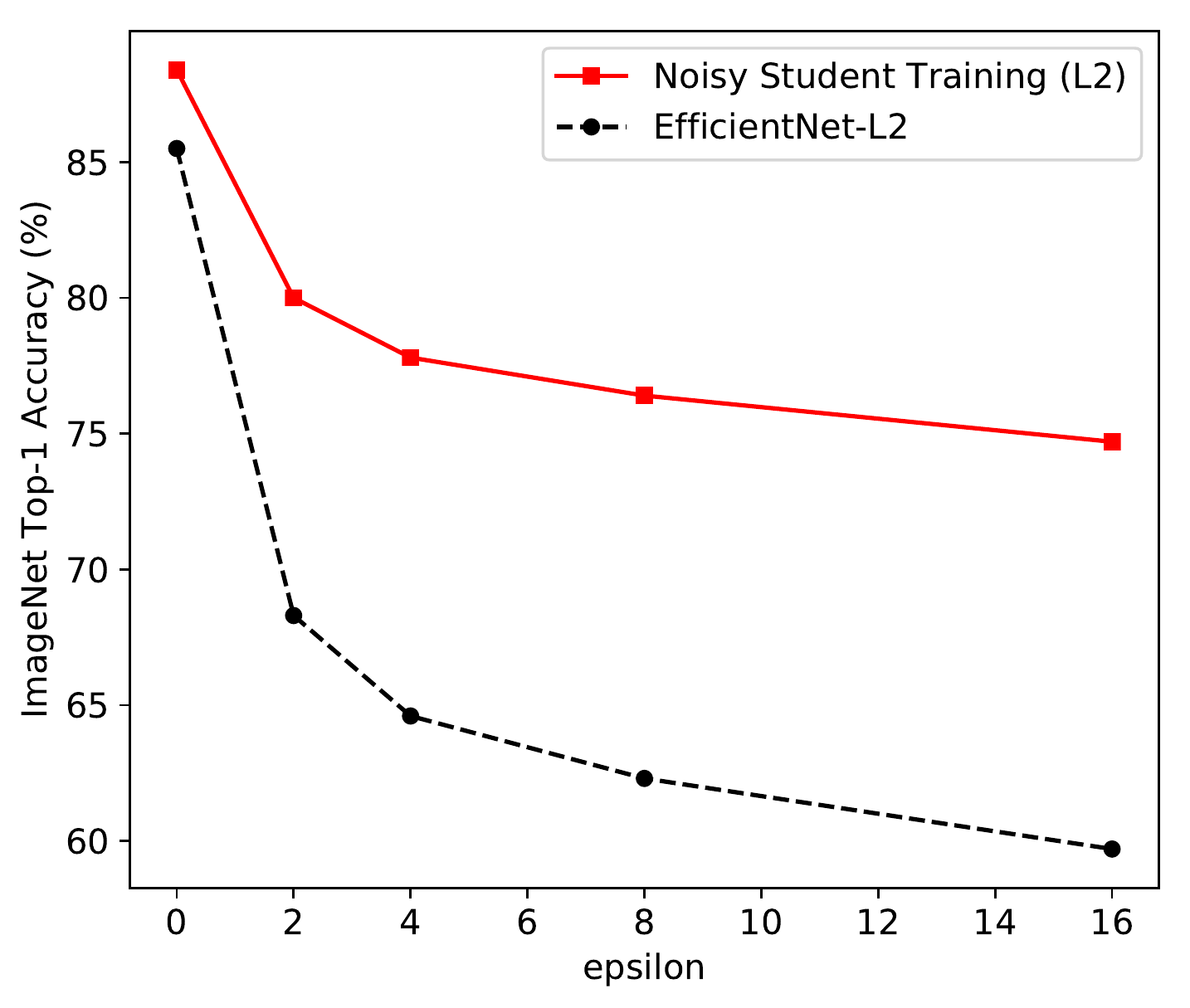}
    \caption{Noisy Student Training improves adversarial robustness against an FGSM attack though the model is not optimized for adversarial robustness. The accuracy is improved by 11\% at $\epsilon = 2$ and gets better as $\epsilon$ gets larger.  }
    \label{fig:adv_robustness}
\end{figure}

After testing our model's robustness to common corruptions and perturbations, we also study its performance on adversarial perturbations. We evaluate our EfficientNet-L2 models with and without Noisy Student Training against an FGSM attack. This attack performs one gradient descent step on the input image~\cite{goodfellow2014explaining} with the update on each pixel set to $\epsilon$. 
As shown in Figure \ref{fig:adv_robustness}, Noisy Student Training leads to very significant improvements in accuracy even though the model is not optimized for adversarial robustness. Under a stronger attack PGD with 10 iterations~\cite{madry2017towards}, at $\epsilon=16$,
Noisy Student Training improves EfficientNet-L2's accuracy from 1.1\% to 4.4\%.

Note that these adversarial robustness results are not directly comparable to prior works since we use a large input resolution of 800x800 and adversarial vulnerability can scale with the input dimension~\cite{galloway2019batch,goodfellow2014explaining,gilmer2018adversarial,simon2019first}.

\section{Ablation Study} 
In this section, we study the importance of noise and iterative training and summarize the ablations for other components of our method. 
\subsection{The Importance of Noise in Self-training}
\label{sec:ablation_noise}
Since we use soft pseudo labels generated from the teacher model, when the student is trained to be exactly the same as the teacher model, the cross entropy loss on unlabeled data would be zero and the training signal would vanish. Hence, a question that naturally arises is why the student can outperform the teacher with soft pseudo labels. As stated earlier, we hypothesize that noising the student is needed so that it does not merely learn the teacher's knowledge. 
We investigate the importance of noising in two scenarios with  different amounts of unlabeled data and different teacher model accuracies. In both cases, we gradually remove augmentation, stochastic depth and dropout for unlabeled images when training the student model, while keeping them for labeled images. 
This way, we can isolate the influence of noising on unlabeled images from the influence of preventing overfitting for labeled images. In addition, we compare using a noised teacher and an unnoised teacher to study if it is necessary to disable noise when generating pseudo labels.

Here, we show the evidence in Table \ref{tab:abl_noise}, noise such as stochastic depth, dropout and data augmentation plays an important role in enabling the student model to perform better than the teacher. The performance consistently drops with noise function removed. 
However, in the case with 130M  unlabeled images, when compared to the supervised baseline, the performance is still improved to 84.3\% from 84.0\% with noise function removed. We hypothesize that the improvement can be attributed to SGD, which introduces stochasticity into the training process. 

\begin{table}[h!]
\small
    \centering                                                                 
        \begin{tabular}{l|cc}                                                
        \toprule 
        Model / Unlabeled Set Size &    1.3M & 130M  \\
        \midrule      
	    EfficientNet-B5  & 83.3\% & 84.0\% \\
	    \midrule
		Noisy Student Training (B5) & \bf 83.9\% & \bf 85.1\% \\
		\  student w/o Aug & 83.6\% & 84.6\% \\
		\  student w/o Aug, SD, Dropout & 83.2\% & 84.3\% \\
		\  teacher w. Aug, SD, Dropout  & 83.7\% & 84.4\% \\
        \bottomrule
        \end{tabular}
    \caption{Ablation study of noising. We use EfficientNet-B5 as the teacher model and study two cases with different numbers of unlabeled images and different augmentations. For the experiment with 1.3M unlabeled images, we use the standard augmentation including random translation and flipping for both the teacher and the student. For the experiment with 130M unlabeled images, we use RandAugment.  Aug and SD denote data augmentation and stochastic depth respectively. We remove the noise for unlabeled images while keeping them for labeled images. Here, iterative training is not used and unlabeled batch size is set to be the same as the labeled batch size to save training time.}
    \label{tab:abl_noise}
\end{table}

One might argue that the improvements from using noise can be resulted from preventing overfitting the pseudo labels on the unlabeled images.
We verify that this is not the case when we use 130M unlabeled images since the model does not overfit the unlabeled set from the training loss. While removing noise  leads to a much lower training loss for labeled images, we observe that, for unlabeled images, removing noise leads to a smaller drop in training loss. This is probably because it is harder to overfit the large unlabeled dataset.

Lastly, adding noise to the teacher model that generates pseudo labels leads to lower accuracy, which shows the importance of having a powerful unnoised teacher model.

\subsection{A Study of Iterative Training}
\label{sec:iterative_training}
Here, we show the detailed effects of iterative training. As mentioned in Section \ref{sec:exp_details}, we first train an EfficientNet-B7 model on labeled data and then use
it as the teacher to train an EfficientNet-L2 student model. Then, we iterate this process by putting back the new student model as the teacher model.

As shown in Table \ref{tab:iterative_training}, the model
performance improves to 87.6\% in the first iteration and then to 88.1\% in the second iteration with the same hyperparameters (except using a teacher model with better performance). These results indicate that iterative training is effective in producing increasingly better models. For the last iteration, we make use of a larger ratio between unlabeled batch size and labeled batch size to boost the final performance to 88.4\%.

\begin{table}[h!]
\small
    \centering                                                                 
        \begin{tabular}{ccc|c}                                     
        \toprule 
        Iteration & Model & Batch Size Ratio & Top-1 Acc. \\
        \midrule
        1 & EfficientNet-L2 & 14:1 & 87.6\% \\
        2 & EfficientNet-L2 & 14:1 & 88.1\% \\
        3 & EfficientNet-L2 & 28:1 & 88.4\% \\
        \bottomrule
        \end{tabular}
    \caption{Iterative training improves the accuracy, where batch size ratio denotes the ratio between unlabeled data and labeled data.} 
    \label{tab:iterative_training}
\end{table}

\subsection{Additional Ablation Study Summarization}
We also study the importance of various design choices of Noisy Student Training, hopefully offering a practical guide for readers. With this purpose, we conduct 8 ablation studies in Appendix \ref{sec:ablation_study}. The findings are summarized as follows:
\begin{itemize}[itemsep=0em,topsep=0em] 
    \item \textbf{Finding \#1:} Using \emph{a large teacher model} with better performance leads to better results. 
    \item \textbf{Finding \#2:} \emph{A large amount of unlabeled data} is necessary for better performance.
    \item \textbf{Finding \#3:} \emph{Soft pseudo labels} work better than hard pseudo labels for out-of-domain data in certain cases.
    \item \textbf{Finding \#4:} \emph{A large student model} is important to enable the student to learn a more powerful model.
    \item \textbf{Finding \#5:} \emph{Data balancing} is useful for small models.
    \item \textbf{Finding \#6:}  \emph{Joint training} on labeled data and unlabeled data outperforms the pipeline that first pretrains with unlabeled data and then finetunes on labeled data. 
    \item \textbf{Finding \#7:} Using \emph{a large ratio between unlabeled batch size and labeled batch size} enables models to train longer on unlabeled data to achieve a higher accuracy. 
    \item \textbf{Finding \#8:} \emph{Training the student from scratch} is sometimes better than initializing the student with the teacher and the student initialized with the teacher still requires a large number of training epochs to perform well. 
\end{itemize}

\section{Related works}
\label{sec:related_works}

\paragraph{Self-training.} Our work is based on self-training  (\eg,~\cite{scudder1965probability,yarowsky1995unsupervised,riloff2003learning,riloff1996automatically}). 
Self-training first uses labeled data to train a good teacher model, then use the teacher model to label unlabeled data and finally use the labeled data and unlabeled data to jointly train a student model. 
In typical self-training with the teacher-student framework, noise injection to the student is not used by default, or the role of noise is not fully understood or justified. The main difference between our work and prior works is that we identify the importance of noise, and aggressively inject noise to make the student better.

Self-training was previously used to improve ResNet-50 from 76.4\%  to 81.2\% top-1 accuracy~\cite{billion_large_scale} which is still far from the state-of-the-art accuracy. Yalniz \etal~\cite{billion_large_scale} also  did not show significant improvements in terms of robustness on ImageNet-A, C and P as we did.  In terms of methodology, they proposed to first only train on unlabeled images and then finetune their model on labeled images as the final stage. In Noisy Student Training, we combine these two steps into one because it simplifies the algorithm and leads to better performance in our experiments.

Data Distillation~\cite{radosavovic2018data}, which ensembles predictions for an image with different transformations to strengthen the teacher, is the opposite of our approach of weakening the student. Parthasarathi \etal~\cite{parthasarathi2019lessons} find a small and fast speech recognition model for deployment via knowledge distillation on unlabeled data. As noise is not used and the student is also small, it is difficult to  make the student better than teacher. 
The domain adaptation framework in \cite{Chowdhury19} is related but highly optimized for videos, \eg, prediction on which frame to use in a video. The method in \cite{zhou2018edf} ensembles predictions from multiple teacher models, which is more expensive than our method.

Co-training~\cite{blum1998combining} divides features into two disjoint partitions and trains two models with the two sets of features using labeled data. Their source of ``noise" is the feature partitioning such that two models do not always agree on unlabeled data. Our method of injecting noise to the student model also enables the teacher and the student to make different predictions and is more suitable for ImageNet than partitioning features. 

Self-training / co-training has also been shown to work well for a variety of other tasks including leveraging noisy data~\cite{veit2017learning}, semantic segmentation~\cite{babakhin2019semi}, text classification~\cite{karamanolakis2019leveraging,sun2019learning}. Back translation and self-training have led to significant improvements in machine translation~\cite{sennrich2015improving, edunov2018understanding, he2016dual,cheng2016semi,wu2019exploiting,he2019revisiting}.

\paragraph{Semi-supervised Learning.} 
Apart from self-training, another important line of work in semi-supervised learning~\cite{chapelle2009semi,zhu2005semi} is based on consistency training~\cite{bachman2014learning,rasmus2015semi,laine2016temporal,tarvainen2017mean,miyato2018virtual,luo2018smooth,qiao2018deep,chen2018semi,clark2018semi,park2018adversarial,athiwaratkun2018there,li2019certainty,verma2019interpolation,uda,mixmatch,zhai2019s,lai2019bridging,berthelot2019remixmatch}. 
They constrain model predictions to be invariant to noise injected to the input, hidden states or model parameters. As discussed in Section \ref{sec:method}, consistency regularization works less well on ImageNet because consistency regularization uses a model being trained to generate the pseudo-labels. In the early phase of training, they regularize the model towards high entropy predictions, and prevents it from achieving good accuracy.

Works based on pseudo label~\cite{lee2013pseudo,iscen2019label,shi2018transductive,arazo2019pseudo} are similar to self-training, but also suffer the same problem with consistency training, since they rely on a model being trained instead of a converged model with high accuracy to generate pseudo labels. 
Finally, frameworks in semi-supervised learning also include graph-based methods ~\cite{zhu2003semi,weston2012deep,yang2016revisiting,kipf2016semi}, methods that make use of latent variables as target variables ~\cite{kingma2014semi,maaloe2016auxiliary,yang2017semi} and methods based on low-density separation~\cite{grandvalet2005semi,salimans2016improved,dai2017good}, which might provide complementary benefits to our method.

\paragraph{Knowledge Distillation.} Our work is also related to methods in Knowledge Distillation~\cite{bucilu2006model,ba2014deep,hinton2015distilling,furlanello2018born,balan2015bayesian} via the  use  of soft targets. The main use of knowledge distillation is model compression by making the student model smaller. The main difference between our method and knowledge distillation is that knowledge distillation does not consider unlabeled data and does not aim to improve the student model.

\paragraph{Robustness.} A number of studies, \eg{} \cite{szegedy2013intriguing,hendrycks2018benchmarking,recht2019imagenet,gu2019using}, have shown that vision models lack robustness. 
Addressing the lack of robustness has become an important research direction in machine learning and computer vision in recent years. Our study shows that using unlabeled data improves accuracy and general robustness.
Our finding is consistent with arguments that using unlabeled data can improve \emph{adversarial} robustness~\cite{carmon2019unlabeled,stanforth2019labels,najafi2019robustness,zhai2019adversarially}. 
The main difference between our work and these works is that they directly  optimize adversarial robustness on unlabeled data, whereas we show that Noisy Student Training improves robustness greatly even without directly optimizing robustness. 

\section{Conclusion}
Prior works on weakly-supervised learning  required billions of weakly labeled data to improve state-of-the-art ImageNet models. In this work, we showed that it is possible to use unlabeled images to significantly advance both accuracy and robustness of state-of-the-art ImageNet models. We found that self-training is a simple and effective algorithm to leverage unlabeled data at scale. We improved it by adding noise to the student, hence the name Noisy Student Training, to learn beyond the teacher's knowledge. 

Our experiments showed that Noisy Student Training and EfficientNet can achieve an accuracy of 88.4\% which is 2.9\% higher than without Noisy Student Training. This result is also a new state-of-the-art and 2.0\% better than the previous best method that used an order of magnitude more weakly labeled data~\cite{mahajan2018exploring,touvron2019fixing}.

An important contribution of our work was to show that  Noisy Student Training boosts robustness in computer vision models. Our experiments showed that our model significantly improves performances on ImageNet-A, C and P.

\subsection*{Acknowledgement}
We thank the Google Brain team, Zihang Dai, Jeff Dean, Hieu Pham, Colin Raffel, Ilya Sutskever and Mingxing Tan for insightful discussions, Cihang Xie, Dan Hendrycks and A. Emin Orhan for robustness evaluation, Sergey Ioffe, Guokun Lai, Jiquan Ngiam, Jiateng Xie and Adams Wei Yu for feedbacks on the draft, Yanping Huang, Pankaj Kanwar, Naveen Kumar, Sameer Kumar and Zak Stone for great help with TPUs, Ekin Dogus Cubuk and Barret Zoph for help with RandAugment, Tom Duerig, Victor Gomes, Paul Haahr, Pandu Nayak, David Price, Janel Thamkul, Elizabeth Trumbull, Jake Walker and Wenlei Zhou for help with model releases, Yanan Bao, Zheyun Feng and Daiyi Peng for help with the JFT dataset, Ola Spyra and Olga Wichrowska for help with infrastructure.
{\small
\bibliographystyle{ieee_fullname}
\bibliography{bib}
}

\newpage
\appendix
\section{Experiments}

\subsection{Architecture Details}
\label{sec:architecture_details}
The architecture specifications of EfficientNet-L2 are listed in Table \ref{tab:efficientnet_architectures}. We also list EfficientNet-B7 as a reference. Scaling width and resolution by $c$ leads to an increase factor of $c^2$ in training time and scaling depth by $c$ leads to an increase factor of $c$.
The training time of EfficientNet-L2 is around $5$ times the training time of EfficientNet-B7. 

\begin{table}[h!]
    \centering
    \footnotesize
    \begin{tabular}{l|ccccc}
    \toprule
    Architecture Name & $w$ & $d$  & Train Res. & Test Res. & \# Params \\
    \midrule
    EfficientNet-B7 & 2.0 & 3.1 & 600 & 600 & 66M \\
    EfficientNet-L2& 4.3 & 5.3 & 475 & 800 & 480M \\
    \bottomrule
    \end{tabular}
    \caption{Architecture specifications for EfficientNets used in the paper. The width $w$ and depth $d$ are the scaling factors that need to be contextualized in EfficientNet~\cite{tan2019efficientnet}. Train Res. and Test Res. denote training and testing resolutions respectively. }
    \label{tab:efficientnet_architectures}
\end{table}

\subsection{Ablation Studies}
\label{sec:ablation_study}
In this section, we provide comprehensive studies of various components of our method. 
Since iterative training results in longer training time, we conduct ablation without it. To further save training time, we reduce the training epochs for small models from 700 to 350, starting from Study \#4. We also set the unlabeled batch size to be the same as the labeled batch size for models smaller than EfficientNet-B7 starting from Study \#2.
 
\paragraph{Study \#1: Teacher Model's Capacity.}
Here, we study if using a larger and better teacher model would lead to better results. 
We use our best model Noisy Student Training with EfficientNet-L2, that achieves a top-1 accuracy of 88.4\%, to teach student models with sizes ranging from  EfficientNet-B0 to EfficientNet-B7. We use the standard augmentation instead of RandAugment on unlabeled data in this experiment to give the student model more capacity. This setting is in principle similar to distillation on unlabeled data. 

The comparison is shown in Table \ref{tab:vary_model_size}. Using Noisy Student Training (EfficientNet-L2) as the teacher leads to another 0.5\% to 1.6\%
improvement on top of the improved results by using the same model as the teacher. For example, we can train a medium-sized model EfficientNet-B4, which has fewer parameters than ResNet-50, to an accuracy of 85.3\%. Therefore, \emph{using a large teacher model with better performance leads to better results.}

\begin{table}[h!]
    \centering
    \footnotesize
    \begin{tabular}{l|c|cc}
    \toprule
         Model & \# Params & Top-1 Acc. & Top-5 Acc. \\
         \midrule 
         EfficientNet-B0 & \multirow{3}{*}{5.3M} & 77.3\% & 93.4\%  \\
         Noisy Student Training (B0) & & 78.1\%  & 94.2\% \\
         \bf Noisy Student Training (B0, L2) & & \bf 78.8\%  & \bf 94.5\% \\
         \midrule
         EfficientNet-B1 & \multirow{3}{*}{7.8M} & 79.2\% & 94.4\%  \\
         Noisy Student Training (B1) & & 80.2\% & 95.2\% \\
         \bf Noisy Student Training (B1, L2) & & \bf 81.5\% & \bf 95.8\% \\
         \midrule
         EfficientNet-B2 & \multirow{3}{*}{9.2M} & 80.0\% & 94.9\% \\
         Noisy Student Training (B2) & & 81.1\% & 95.5\% \\
         \bf Noisy Student Training (B2,  L2) & & \bf 82.4\% & \bf 96.3\% \\
         \midrule
         EfficientNet-B3 & \multirow{3}{*}{12M} & 81.7\% & 95.7\%  \\
         Noisy Student Training (B3) & & 82.5\% & 96.4\% \\
         \bf Noisy Student Training (B3, L2) & & \bf 84.1\%  & \bf 96.9\% \\
         \midrule
         EfficientNet-B4 & \multirow{3}{*}{19M} & 83.2\% &  96.4\% \\
         Noisy Student Training (B4) & & 84.4\% & 97.0\%  \\
         \bf Noisy Student Training (B4, L2) & & \bf 85.3\% & \bf 97.5\% \\
         \midrule
         EfficientNet-B5 & \multirow{3}{*}{30M} & 84.0\% & 96.8\%  \\
         Noisy Student Training (B5) & & 85.1\%  & 97.3\% \\
         \bf Noisy Student Training (B5, L2) & & \bf 86.1\% & \bf 97.8\% \\
         \midrule
         EfficientNet-B6 & \multirow{3}{*}{43M} & 84.5\% & 97.0\%  \\
         Noisy Student Training (B6) & & 85.9\% & 97.6\% \\
         \bf Noisy Student Training (B6, L2) & & \bf 86.4\% & \bf 97.9\% \\
         \midrule
         EfficientNet-B7 & \multirow{3}{*}{66M} & 85.0\% & 97.2\%  \\
         Noisy Student Training (B7) & & 86.4\% & 97.9\% \\
         \bf Noisy Student Training (B7, L2) & & \bf 86.9\% & \bf 98.1\% \\
         \bottomrule
    \end{tabular}
    \caption{Using our best model with 88.4\% accuracy as the teacher (denoted as Noisy Student Training (X, L2)) leads to more improvements than using the same model as the teacher (denoted as Noisy Student Training (X)).
    Models smaller than EfficientNet-B5 are trained for 700 epochs (better than training for 350 epochs as used in Study \#4 to Study \#8).
     Models other than EfficientNet-B0 uses an unlabeled batch size of three times the labeled batch size, while other ablation studies set the unlabeled batch size to be the same as labeled batch size by default for models smaller than B7. }
    \label{tab:vary_model_size}
\end{table}

\paragraph{Study \#2: Unlabeled Data Size.}
Next, we conduct experiments to understand the effects of using different amounts of unlabeled data. We start with the 130M unlabeled images and gradually reduce the unlabeled set. We experiment with using $\frac{1}{128}, \frac{1}{64}, \frac{1}{32}, \frac{1}{16}, \frac{1}{4}$ of the whole data by uniformly sampling images from the the unlabeled set for simplicity, though taking images with highest confidence may lead to better results. We use EfficientNet-B4 as both the teacher and the student. 

As can be seen from Table \ref{tab:vary_unlabeled_data}, the performance stays similar when we reduce the data to $\frac{1}{16}$ of the whole data,\footnote{A larger model might benefit from more data while a small model with limited capacity can easily saturate.}  which amounts to 8.1M images after duplicating. The performance drops when we further reduce it. Hence, \emph{using a large amount of unlabeled data leads to better performance}.

\begin{table}[h!]
\footnotesize
    \centering
    \begin{tabular}{l|cccccc}
    \toprule
    Data & $1/128$ & $1/64$ & $1/32$ & $1/16$ & $1/4$ & $1$ \\
    \midrule
    Top-1 Acc. &83.4\% & 83.3\% & 83.7\% & 83.9\% & 83.8\% & \bf 84.0\% \\
    \bottomrule
    \end{tabular}
    \caption{Noisy Student Training's performance improves with more unlabeled data. Models are trained for 700 epochs without iterative training. The baseline model achieves an accuracy of 83.2\%.}
    \label{tab:vary_unlabeled_data}
\end{table}

\paragraph{Study \#3: Hard Pseudo-Label vs. Soft Pseudo-Label on Out-of-domain Data.}

Unlike previous studies in semi-supervised learning that use in-domain unlabeled data (\eg, CIFAR-10 images as unlabeled data for a small CIFAR-10 training set), to improve ImageNet, we must use  out-of-domain unlabeled data. 
Here we compare hard pseudo-label and soft pseudo-label for out-of-domain data. 
Since a teacher model's confidence on an image can be a good indicator of whether it is an out-of-domain image, we consider the high-confidence images as in-domain images and the low-confidence images as out-of-domain images. We sample 1.3M images in each confidence interval $[0.0, 0.1], [0.1, 0.2], \cdots, [0.9, 1.0]$.

 \begin{figure}[h!]
    \centering     
    \includegraphics[width=0.85\columnwidth]{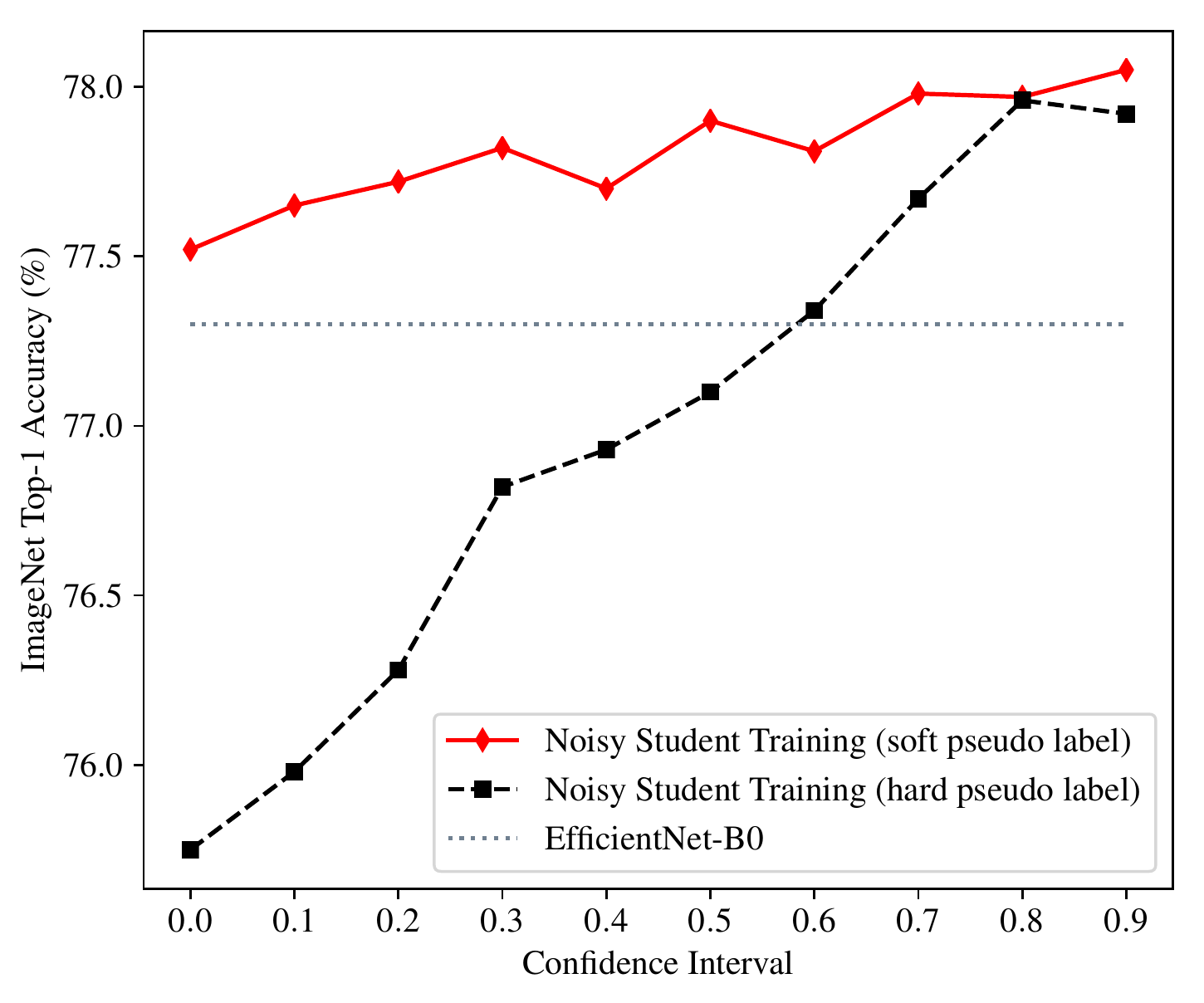}     
    \caption{Soft pseudo labels lead to better performance for low confidence data (out-of-domain data). Each dot at $p$ represents a Noisy Student Training model trained with 1.3M ImageNet labeled images and 1.3M unlabeled images with confidence scores in $[p, p+0.1]$.}  
    \label{fig:soft_vs_hard_vary_confidence} 
 \end{figure}
We use EfficientNet-B0 as both the teacher model and the student model and compare using Noisy Student Training with soft pseudo labels and hard pseudo labels. 
The results are shown in Figure  \ref{fig:soft_vs_hard_vary_confidence} with the following observations: \emph{(1) Soft pseudo labels and hard pseudo labels can both lead to significant improvements with in-domain unlabeled images \ie, high-confidence images. (2) With out-of-domain unlabeled images, hard pseudo labels can hurt the performance while soft pseudo labels lead to robust performance.}

Note that we have also observed that using hard pseudo labels can achieve as good results or slightly better results when a larger teacher is employed. Hence, whether soft pseudo labels or hard pseudo labels work better might need to be determined on a case-by-case basis.

\paragraph{Study \#4: Student Model's Capacity.} Then, we investigate the effects of student models with different capacities. For teacher models, we use EfficientNet-B0, B2 and B4 trained on labeled data and EfficientNet-B7 trained using Noisy Student Training. We compare using a student model with the same size or with a larger size. The comparison is shown in Table  \ref{tab:abl_student_model}. With the same teacher, using a larger student model leads to consistently better performance, showing that \emph{using a large student model is important to enable the student to learn a more powerful model.}
\begin{table}[h!]
\footnotesize
    \centering
        \begin{tabular}{cc|cc}
        \toprule 
         Teacher & Teacher Acc. & Student  & Student Acc.  \\
        \midrule      
	    \multirow{2}{*}{B0} & \multirow{2}{*}{77.3\%} & B0 & 77.9\% \\
	     & & B1 & \bf 79.5\% \\
	    \midrule
	    \multirow{2}{*}{B2} & \multirow{2}{*}{80.0\%}  & B2 & 80.7\% \\
	     & & B3 & \bf 82.0\% \\
	    \midrule
	    \multirow{2}{*}{B4} & \multirow{2}{*}{83.2\%} & B4 & 84.0\% \\
	     & & B5 & \bf 84.7\% \\
	     \midrule
	    \multirow{2}{*}{B7} & \multirow{2}{*}{86.9\%} & B7 & 86.9\% \\
	     & & L2 & \bf 87.2\% \\
        \bottomrule
        \end{tabular}
    \caption{Using a larger student model leads to better performance. Student models are trained for 350 epochs instead of 700 epochs without iterative training. The B7 teacher with an accuracy of 86.9\% is trained by Noisy Student Training with multiple iterations using B7. The comparison between B7 and L2 as student models is not completely fair for L2, since we use an unlabeled batch size of 3x the labeled batch size for training L2, which is not as good as using an unlabeled batch size of 7x the labeled batch size when training B7 (See Study \#7 for more details). }
    \label{tab:abl_student_model}
\end{table}

\paragraph{Study \#5: Data Balancing.}
Here, we study the necessity of keeping the unlabeled data balanced across categories. As a comparison, we use all unlabeled data that has a confidence score higher than 0.3.  We present results with EfficientNet-B0 to B3 as the backbone models in Table \ref{tab:abl_data_balancing}. Using data balancing leads to better performance for small models EfficientNet-B0 and B1. Interestingly, the gap becomes smaller for larger models such as EfficientNet-B2 and B3, which shows that more powerful models can learn from unbalanced data effectively. 
\emph{To enable Noisy Student Training to work well for all model sizes, we use data balancing by default.}
\begin{table}[h!]
\footnotesize
    \centering
        \begin{tabular}{l|cccc}
        \toprule 
        Model &  B0 & B1 & B2 & B3  \\
        \midrule      
	    Supervised Learning  & 77.3\% & 79.2\% & 80.0\% & 81.7\% \\
	    \midrule
		Noisy Student Training & \bf 77.9\% & \bf 79.9\% & \bf 80.7\% & 82.1\% \\
		\  w/o Data Balancing &  77.6\% & 79.6\% & 80.6\% &  82.1\% \\
        \bottomrule
        \end{tabular}
    \caption{Data balancing leads to better results for small models. Models are trained for 350 epochs instead of 700 epochs without iterative training.}
    \label{tab:abl_data_balancing}
\end{table}

\paragraph{Study \#6: Joint Training.} 
In our algorithm, we train the model with labeled images and pseudo-labeled images jointly. Here, we also compare with an alternative approach used by Yalniz \etal \cite{billion_large_scale}, which first pretrains the model on pseudo-labeled images and then finetunes it on labeled images. For finetuning, we experiment with different steps and take the best results. The comparison is shown in Table \ref{tab:abl_joint_training}. 

It is clear that joint training significantly outperforms pretraining + finetuning. 
Note that pretraining only on pseudo-labeled images leads to a much lower accuracy than supervised learning only on labeled data, which suggests that the distribution of unlabeled data is very different from that of labeled data. \emph{In this case, joint training leads to a better solution that fits both types of data.}

\begin{table}[h!]
\footnotesize
    \centering
        \begin{tabular}{l|cccc}
        \toprule 
        Model &  B0 & B1 & B2 & B3  \\
        \midrule      
	    Supervised Learning  & 77.3\% & 79.2\% & 80.0\% & 81.7\% \\
	    \midrule
		Pretraining              & 72.6\% & 75.1\% & 75.9\% & 76.5\% \\
		Pretraining + Finetuning & 77.5\% & 79.4\% & 80.3\% & 81.7\% \\
		Joint Training           & \bf 77.9\% & \bf 79.9\% & \bf 80.7\% & \bf 82.1\% \\
        \bottomrule
        \end{tabular}
    \caption{Joint training works better than pretraining and finetuning.  We vary the finetuning steps and report the best results. Models are trained for 350 epochs instead of 700 epochs without iterative training. }
    \label{tab:abl_joint_training}
\end{table}

\paragraph{Study \#7: Ratio between Unlabeled Batch Size and Labeled Batch Size.} 
Since we use 130M unlabeled images and 1.3M labeled images, if  the batch sizes for unlabeled data and labeled data are the same, the model is trained on unlabeled data only for one epoch every time it is trained on labeled data for a hundred epochs. Ideally, we would also like the model to be trained on unlabeled data for more epochs by using a larger unlabeled batch size so that it can fit the unlabeled data better. Hence we study the importance of the ratio between unlabeled batch size and labeled batch size.

\begin{table}[h!]
\footnotesize
    \centering
        \begin{tabular}{l|cc}
        \toprule 
         Teacher (Acc.) & Batch Size Ratio  & Top-1 Acc.  \\
        \midrule      
        \multirow{2}{*}{B4 (83.2)} & 1:1 & 84.0\% \\
	     & 3:1 & 84.0\% \\
	     \midrule      
	    \multirow{2}{*}{L2 (87.0)} & 1:1 & 86.7\% \\
	     & 3:1 & \bf 87.4\% \\
	    \midrule
	    \multirow{2}{*}{L2 (87.4)} & 3:1 & 87.4\% \\
	     & 6:1 & \bf 87.9\% \\
        \bottomrule
        \end{tabular}
    \caption{With a fixed labeled batch size, a larger unlabeled batch size leads to better performance for EfficientNet-L2. The Batch Size Ratio denotes the ratio between unlabeled batch size and labeled batch size.}
    \label{tab:abl_unlabeled_ratio}
\end{table}

In this study, we try a medium-sized model EfficientNet-B4 as well as a larger model EfficientNet-L2. We use models of the same size as both the teacher and the student. As shown in Table \ref{tab:abl_unlabeled_ratio}, the larger model EfficientNet-L2 benefits from a large ratio while the smaller model EfficientNet-B4 does not. \emph{Using a larger ratio between unlabeled batch size and labeled batch size, leads to substantially better performance for a large model. }

\paragraph{Study \#8: Warm-starting the Student Model.}
Lastly, one might wonder if we should train the student model from scratch when it can be initialized with a converged teacher model with good accuracy. In this ablation, we first train an EfficientNet-B0 model on ImageNet and use it to initialize the student model. We vary the number of epochs for training the student and use the same exponential decay learning rate schedule. Training starts at different learning rates so that the learning rate is decayed to the same value in all experiments. As shown in Table \ref{tab:warm_start}, the accuracy drops significantly when we reduce the training epoch from 350 to 70 and drops slightly when reduced to 280 or 140. Hence, the student still needs to be trained for a large number of epochs even with warm-starting.

Further, we also observe that a student initialized with the teacher can sometimes be stuck in a local optimal. For example, when we use EfficientNet-B7 with an accuracy of 86.4\% as the teacher, the student model initialized with the teacher achieves an accuracy of 86.4\% halfway through the training but gets stuck there when trained for 210 epochs, while a model trained from scratch achieves an accuracy of 86.9\%. Hence, though we can save training time by warm-staring, \emph{we train our model from scratch to ensure the best performance.}

\begin{table}[h!]
\footnotesize
    \centering
    \begin{tabular}{l|cccc|c}
    \toprule
    Warm-start & \multicolumn{4}{c|}{Initializing student with teacher} & No Init \\
    Epoch & 35 & 70 & 140 & 280 & 350 \\
    \midrule
    Top-1 Acc. & 77.4\%  & 77.5\% & 77.7\% & 77.8\% & \bf 77.9\% \\
    \bottomrule
    \end{tabular}
    \caption{A student initialized with the teacher still requires at least 140 epochs to perform well. The baseline model, trained with labeled data only, has an accuracy of 77.3\%.}
    \label{tab:warm_start}
\end{table}

\subsection{Results with a Different Architecture and Dataset}
\paragraph{Results with ResNet-50.} To study whether other architectures can benefit from Noisy Student Training, we conduct experiments with ResNet-50~\cite{he2016deep}. We use the full ImageNet as the labeled data and the 130M images from JFT as the unlabeled data. We train a ResNet-50 model on ImageNet and use it as our teacher model. We use RandAugment with the magnitude set to 9 as the noise. 

The results are shown in Table \ref{tab:results_resnet_50}. Noisy Student Training leads to an improvement of 1.3\% on the baseline model, which shows that Noisy Student Training is effective for architectures other than EfficientNet.

\begin{table}[h!]
\footnotesize
    \centering
        \begin{tabular}{l|cc}
        \toprule 
        Method &  Top-1 Acc. & Top-5 Acc.  \\
        \midrule      
	    ResNet-50  & 77.6\% & 93.8\% \\
		\bf Noisy Student Training (ResNet-50) & \bf 78.9\% & \bf 94.3\%  \\
        \bottomrule
        \end{tabular}
    \caption{Experiments on ResNet-50.}
    \label{tab:results_resnet_50}
\end{table}

\paragraph{Results on SVHN.} We also evaluate Noisy Student Training on a smaller dataset SVHN. We use the core set with 73K images as the training set and the validation set. The extra set with 531K images are used as the unlabeled set. We use EfficientNet-B0 with strides of the second and the third blocks set to 1 so that the final feature map is 4x4 when the input image size is 32x32. 

As shown in Table \ref{tab:results_svhn}, Noisy Student Training improves the baseline accuracy from 98.1\% to 98.6\% and outperforms the previous state-of-the-art results achieved by RandAugment with Wide-ResNet-28-10.

\begin{table}[h!]
\footnotesize
    \centering
        \begin{tabular}{l|c}
        \toprule 
        Method & Accuracy \\
        \midrule
        RandAugment (WRN) & 98.3\% \\
        \midrule
        RandAugment (EfficientNet-B0) & 98.1\% \\ 
        \bf Noisy Student Training (B0) & \bf 98.6\% \\
        \bottomrule
        \end{tabular}
    \caption{Results on SVHN.}
    \label{tab:results_svhn}
\end{table}

\subsection{Results on YFCC100M}
\label{sec:exp_yfcc100m}

\begin{table}[h!]
    \centering
    \footnotesize
    \begin{tabular}{l|c|cc}
    \toprule
         Model & Dataset & Top-1 Acc. & Top-5 Acc. \\
         \midrule 
         EfficientNet-B0 & - & 77.3\% & 93.4\%  \\
         Noisy Student Training (B0) & YFCC & 79.9\%  & 95.0\% \\
         \bf Noisy Student Training (B0) & JFT & \bf 78.1\%  & \bf 94.2\% \\
         \midrule
         EfficientNet-B1 & - & 79.2\% & 94.4\%  \\
         Noisy Student Training (B1) & YFCC & 79.9\% & 95.0\% \\
         \bf Noisy Student Training (B1) & JFT & \bf 80.2\% & \bf 95.2\% \\
         \midrule
         EfficientNet-B2 & - & 80.0\% & 94.9\% \\
         Noisy Student Training (B2) & YFCC & 81.0\% & \bf 95.6\% \\
         \bf Noisy Student Training (B2) & JFT & \bf 81.1\% & 95.5\% \\
         \midrule
         EfficientNet-B3 & - & 81.7\% & 95.7\%  \\
         Noisy Student Training (B3) & YFCC & 82.3\% & 96.2\% \\
         \bf Noisy Student Training (B3) & JFT & \bf 82.5\% & \bf 96.4\% \\
         \midrule
         EfficientNet-B4 & - & 83.2\% &  96.4\% \\
         Noisy Student Training (B4) & YFCC & 84.2\% & 96.9\%  \\
         \bf Noisy Student Training (B4) & JFT & \bf 84.4\% & \bf 97.0\%  \\
         \midrule
         EfficientNet-B5 & - & 84.0\% & 96.8\%  \\
         Noisy Student Training (B5) & YFCC &  85.0\%  & 97.2\% \\
         \bf Noisy Student Training (B5) & JFT & \bf 85.1\%  &  \bf 97.3\% \\
         \midrule
         EfficientNet-B6 & - & 84.5\% & 97.0\%  \\
         Noisy Student Training (B6) & YFCC & 85.4\% & 97.5\% \\
         \bf Noisy Student Training (B6) & JFT & \bf 85.6\% & \bf 97.6\% \\
         \midrule
         EfficientNet-B7 & - & 85.0\% & 97.2\%  \\
         Noisy Student Training (B7) & YFCC &  86.2\% &  \bf 97.9\% \\
         \bf Noisy Student Training (B7) & JFT &  \bf 86.4\% & \bf 97.9\% \\
         \bottomrule
    \end{tabular}
    \caption{Results using YFCC100M and JFT as the unlabeled dataset. }
    \label{tab:results_yfcc}
\end{table}
Since JFT is not a public dataset, we also experiment with a public unlabeled dataset YFCC100M~\cite{thomee2016yfcc100m}, so that researchers can make fair comparisons with our results. Similar to the setting in Section \ref{sec:imagenet_results}, we experiment with different model sizes without iterative training. We use the same model for both the teacher and the student. We also use the same hyperparamters when using JFT and YFCC100M.
Similar to the case for JFT, we first filter images from ImageNet validation set. We then filter low confidence images according to B0's prediction and only keep the top 130K images for each class according to the top-1 predicted class. The resulting set has 34M images since there are not enough images for most classes. We then balance the dataset and increase it to 130M images. As a comparison, before the data balancing stage, there are 81M images in JFT.

As shown in Table \ref{tab:results_yfcc}, Noisy Student Training also leads to significant improvements using YFCC100M though it achieves better performance using JFT. 
The performance difference is probably due to the dataset size difference.

\subsection{Details of Robustness Benchmarks}
\label{sec:robustness}
\paragraph{Metrics.} For completeness, we provide brief descriptions of metrics used in robustness benchmarks ImageNet-A, ImageNet-C and ImageNet-P.
\begin{itemize}
    \item \textbf{ImageNet-A.} The top-1 and top-5 accuracy are measured on the 200 classes that ImageNet-A includes. The mapping from the 200 classes to the original ImageNet classes are available online.\footnote{https://github.com/hendrycks/natural-adv-examples/blob/master/eval.py}
    \item \textbf{ImageNet-C.} mCE (mean corruption error) is the weighted average of error rate on different corruptions, with AlexNet's error rate as a baseline. The score is normalized by AlexNet's error rate so that corruptions with different difficulties lead to scores of a similar scale. Please refer to ~\cite{hendrycks2018benchmarking} for details about mCE and AlexNet's error rate. The top-1 accuracy is simply the average top-1 accuracy for all corruptions and all severity degrees. The top-1 accuracy of prior methods are computed from their reported corruption error on each corruption.
\item \textbf{ImageNet-P.} Flip probability is the probability that the model changes top-1 prediction for different perturbations. mFR (mean flip rate) is the weighted average of flip probability on different perturbations, with AlexNet's flip probability as a baseline. Please refer to ~\cite{hendrycks2018benchmarking} for details about mFR and AlexNet's flip probability. The top-1 accuracy reported in this paper is the average accuracy for all images included in ImageNet-P.
\end{itemize}

\paragraph{On Using RandAugment for ImageNet-C and ImageNet-P.} 
Since Noisy Student Training leads to significant  improvements on ImageNet-C and ImageNet-P, we briefly discuss the influence of RandAugment on robustness results. First, note that our supervised baseline EfficientNet-L2 also uses RandAugment. Noisy Student Training leads to significant improvements when compared to the supervised baseline as shown in Table \ref{tab:robustness2} and Table \ref{tab:robustness3}. 

Second, the overlap between transformations of RandAugment and ImageNet-C, P is small. For completeness, we list transformations in RandAugment and corruptions and perturbations in ImageNet-C and ImageNet-P here:
\begin{itemize}
    \item RandAugment transformations: AutoContrast, Equalize, Invert, Rotate, Posterize, Solarize, Color, Contrast, Brightness, Sharpness, ShearX, ShearY, TranslateX and TranslateY.
    \item Corruptions in ImageNet-C: Gaussian Noise, Shot Noise, Impulse Noise, Defocus Blur, Frosted Glass Blur, Motion Blur, Zoom Blur, Snow, Frost, Fog, Brightness, Contrast, Elastic, Pixelate, JPEG.
    \item Perturbations in ImageNet-P: Gaussian Noise, Shot Noise, Motion Blur, Zoom Blur, Snow, Brightness, Translate, Rotate, Tilt, Scale.
\end{itemize} 

The main overlap between RandAugment and ImageNet-C are Contrast, Brightness and Sharpness. 
Among them, augmentation Contrast and Brightness are also used in ResNeXt-101 WSL \cite{mahajan2018exploring} and in vision models that uses the Inception preprocessing~\cite{howard2013some,szegedy2015going}. 
The overlap between RandAugment and ImageNet-P includes Brightness, Translate and Rotate. 

\end{document}